%% file: acl2021.tex
\documentclass[11pt,a4paper]{article}
\usepackage[hyperref]{acl2021}
\usepackage{url}
\usepackage{latexsym}

\usepackage{times}
\usepackage{amsmath}
\usepackage{bm}
\usepackage{placeins}
\usepackage{booktabs}
\usepackage{amsfonts}
\usepackage{graphicx}
\usepackage{enumitem}
\usepackage{array,multirow,graphicx}
\usepackage{float}
\usepackage{bbm}
\usepackage{mathtools, nccmath}
\usepackage{microtype}

\newcommand{\specialcell}[2][c]{\begin{tabular}[#1]{@{}c@{}}#2\end{tabular}}
\newcommand{\norm}[1]{\left\lVert#1\right\rVert}

\aclfinalcopy 

\title{A Systematic Investigation of KB-Text Embedding Alignment at Scale}

\author{
Vardaan Pahuja\textsuperscript{1},\quad 
Yu Gu\textsuperscript{1},\quad
Wenhu Chen\textsuperscript{2},\quad 
Mehdi Bahrami\textsuperscript{3},\\
\textbf{Lei Liu\textsuperscript{3},}\quad
\textbf{Wei-Peng Chen\textsuperscript{3},}\quad
\textbf{Yu Su\textsuperscript{1}}\\
\textsuperscript{1}The Ohio State University\quad\quad
\textsuperscript{2}University of California, Santa Barbara\quad\quad\\
\textsuperscript{3}Fujitsu Laboratories of America\\
{\small\tt \{pahuja.9, gu.826, su.809\}@osu.edu},\quad
{\small\tt wenhuchen@cs.ucsb.edu}\\
{\small\tt \{mbahrami, lliu, wchen\}@fujitsu.com}\\
}

\date{}

\begin{document}
\maketitle
\begin{abstract}


Knowledge bases (KBs) and text often contain complementary knowledge: KBs store structured knowledge that can support long-range reasoning, while text stores more comprehensive and timely knowledge in an unstructured way. 
Separately embedding the individual knowledge sources into vector spaces has demonstrated tremendous successes in encoding the respective knowledge, but how to jointly embed and reason with both knowledge sources to fully leverage the complementary information is still largely an open problem.
We conduct a large-scale, systematic investigation of aligning KB and text embeddings for joint reasoning. 
We set up a novel evaluation framework with two evaluation tasks, \textit{few-shot link prediction} and \textit{analogical reasoning}, and evaluate an array of KB-text embedding alignment methods. We also demonstrate how such alignment can infuse textual information into KB embeddings for more accurate link prediction on emerging entities and events, using COVID-19 as a case study.\footnote{Code and data are available at \url{https://github.com/dki-lab/joint-kb-text-embedding}.}

\end{abstract}

\section{Introduction}
\input{intro.tex}

\section{Related Work}
\input{related_work.tex}

\section{Model}
\input{model.tex}

\section{Dataset}
\input{dataset.tex}

\section{Experiments}

\input{experiments.tex}

\section{Conclusion}
\input{conclusion.tex}

\section*{Acknowledgements}
\input{acknowledgements}

\FloatBarrier




\bibliographystyle{acl_natbib}
\bibliography{anthology,acl2021}


\end{document}

%% file: intro.tex
Recent years have witnessed a rapid growth of knowledge bases (KBs) such as Freebase \cite{bollacker2007platform}, DBPedia \cite{auer2007dbpedia}, YAGO \cite{suchanek2007yago} and Wikidata \cite{vrandevcic2014wikidata}. These KBs store facts about real-world entities (e.g. people, places, and things) in the form of RDF triples, i.e. (subject, predicate, object). Today’s KBs are massive in scale. For instance, Freebase contains over 45 million entities and 3 billion facts involving a large variety of relations. Such large-scale multi-relational knowledge provides a great potential for improving a wide range of tasks, from information retrieval \cite{Castells2007AnAO, Shen2015EntityLW}, question answering \cite{Yao2014InformationEO, Yu2017ImprovedNR} to biological data mining \cite{Zheng2020PharmKGAD}.

\begin{figure}[!t]
\centering
\resizebox{0.95\linewidth}{!}{
\centering\includegraphics[scale=1.0]{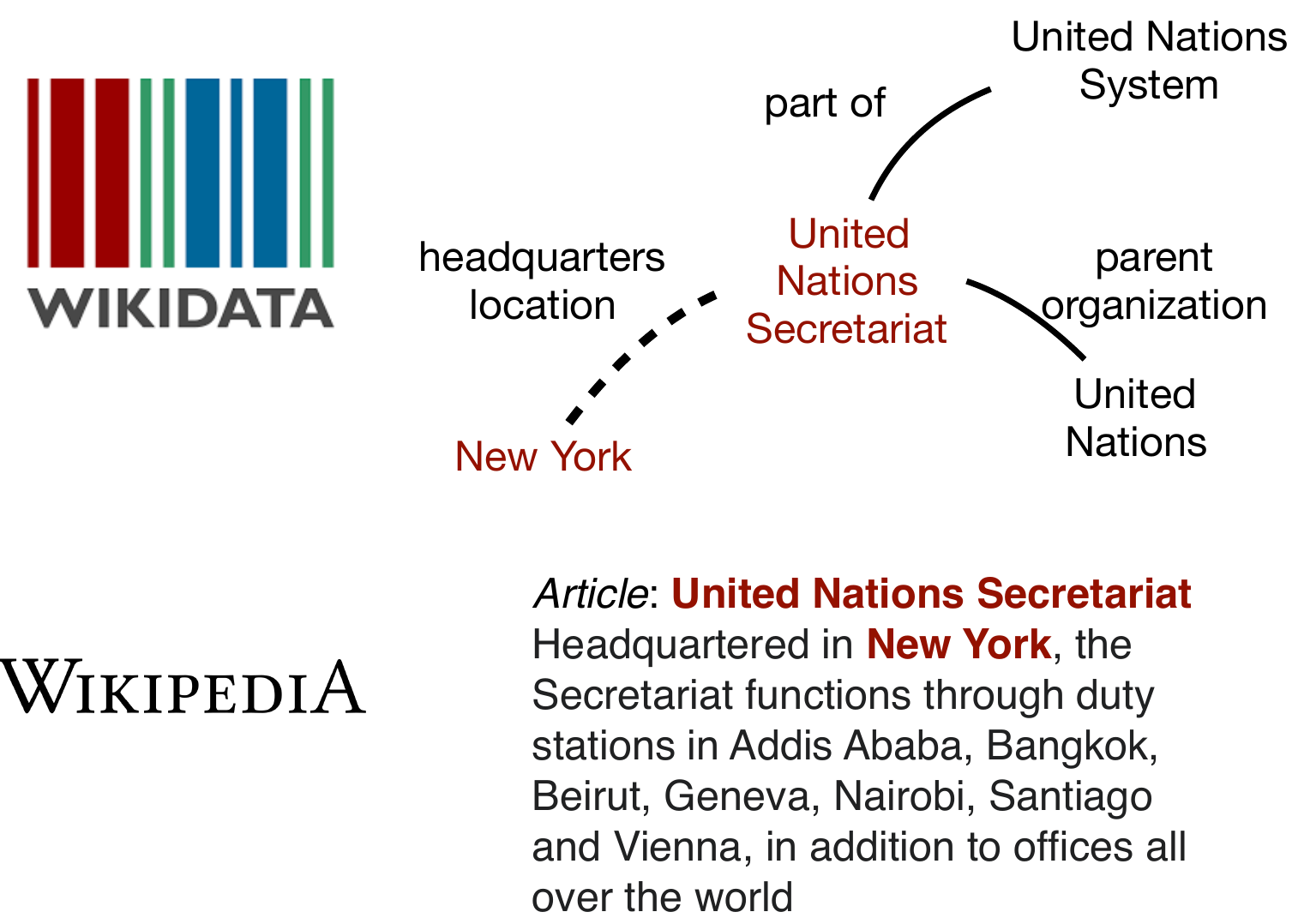}}
\caption{KBs and text are complementary and embedding alignment could help injecting information from one source to the other and vice versa. Dashed line is missing link in the KB.}
\label{fig:front_fig}
\end{figure}

KB embedding models~\cite{bordes2013translating, dong2014knowledge, lin2015learning} embed entities and relations into vector space(s) such that the embeddings capture the symbolic knowledge present in the KB. Similarly, word embedding models \cite{mikolov2013distributed, pennington2014glove} learn continuous vector representations that capture the distributional semantics of words. Experiments on analogical reasoning~\cite{mikolov2013distributed,gladkova2016analogy} and multilingual word embedding alignment \cite{mikolov2013exploiting} have shown that there exists a linear structure in the word embedding space encoding relational information. On the other hand, translation-based KB embedding models \cite{bordes2013translating, lin2015learning, ji2015knowledge}, by construction, also present a linear structure in their embedding space.

A natural question then is, \emph{can we align the two embedding spaces such that they mutually enhance each other?} 
Such alignment could potentially inject structured knowledge from KBs into text embeddings and inject unstructured but more timely-updated knowledge from text into KB embeddings, leading to more universal and comprehensive embeddings (Figure~\ref{fig:front_fig}). Several studies have attempted at this. \citet{lao2012reading} use the Path-Ranking Algorithm \cite{lao2010relational} on combined text and KB to improve binary relation prediction. \citet{gardner2014incorporating} leverage text data to enhance KB inference and help address the incompleteness of KBs. \citet{toutanova2015representing} augment the KB with facts and relations from the text corpus and learn joint embedding for entities, KB relations and textual relations. 
Enhancement of KB entity embeddings using using Entity Descriptions has been attempted in \cite{zhong2015aligning, xie2016representation}. \citet{wang2014knowledge} propose to jointly embed entities and words in the same vector space. The alignment of embeddings of words and entities is accomplished using Wikipedia anchors or entity names.

However, existing studies are still ad-hoc and a more systematic investigation of KB-text embedding alignment is needed to answer an array of important open questions: \emph{What is the best way to align the KB and text embedding spaces?} \emph{To what degree can such alignment inject information from one source to another?} \emph{How to balance the alignment loss with the original embedding losses?} In this work, we conduct a systematic investigation of KB-text embedding alignment at scale and seek to answer these questions. Our investigation uses the latest version of the full Wikidata~\cite{vrandevcic2014wikidata} as the KB, the full Wikipedia as the text corpus, and the shared entities as anchors for alignment. We define two tasks, few-shot link prediction and analogical reasoning, to evaluate the effectiveness of injecting text information into KB embeddings and injecting KB information into text embeddings, respectively, based on which we evaluate and compare an array of embedding alignment methods. The results and discussion present new insights about this important problem. Finally, using COVID-19 as a case study, we also demonstrate that such alignment can effectively inject text information into KB embeddings to complete KBs on emerging entities and events.

In summary, our contributions are three-fold:

\begin{enumerate}
    \item We conduct the first systematic investigation on KB-text embedding alignment at scale and propose and compare multiple effective alignment methods.
    \item We set up a novel evaluation framework with two evaluation tasks, few-shot link prediction and analogical reasoning, to facilitate future research on this important problem.
    \item We have also learned joint KB-text embeddings on the largest-scale data to date and will release the embeddings as a valuable resource to the community.
\end{enumerate}

%% file: related_work.tex
\noindent \textbf{KB-KB embedding alignment.} Most existing knowledge bases are incomplete. Learning of distributed representations for entities and relations in knowledge bases finds application in the task of link prediction i.e. to infer missing facts in the KB given the known facts. This includes translation-based models \cite{bordes2013translating, lin2015learning, ji2015knowledge}, feed-forward neural network based approaches \cite{socher2013reasoning, dong2014knowledge}, convolutional neural networks \cite{dettmers2018convolutional, nguyen2017novel} and models that leverage graph neural networks \cite{schlichtkrull2018modeling, shang2019end,nathani2019learning}.
Recently, many research works have focused on the alignment of embedding spaces of heterogeneous data sources such as different KBs.
JE \cite{hao2016joint} introduces a projection matrix to align the embedding spaces of different KBs. MTransE \cite{chen2016multilingual} first learns the embeddings of entities and relations in each language independently and then learns the transformation between these embedding spaces. \citet{wang2018cross} use Graph Convolutional networks and a set of pre-aligned entities to learn embeddings of entities in multilingual KBs in a unified vector space. In the present work, we focus on aligning the KB and textual embedding spaces.

\noindent \textbf{KB-text joint representation.} Many recent approaches have attempted to learn the embeddings of words and knowledge base entities in the same vector space. 
\citet{wang2014knowledge} propose an alignment technique for KB and text representations using entity names and/or anchors. Wikipedia2Vec \cite{yamada2016joint} extends the skip-gram based model by modeling entity-entity co-occurrences using a link graph and word-entity co-occurrences using KB anchors. 
However, an entity mention can be ambiguous i.e. it can refer to different entities in different contexts. To resolve this, \citet{cao2017bridge} propose Multi-Prototype Entity
Mention Embedding model to learn representations for different senses of entity mentions.
It includes a mention sense embedding model which uses context words and a set of reference entities to predict the actual entity referred to by the mention. Despite this progress, a comprehensive investigation of the merits of different alignment approaches is missing. Our work takes a step forward in this direction and proposes a novel evaluation framework to compare multiple alignment approaches for KB-Text joint embedding on a large-scale KB and textual corpus.

%% file: model.tex
\begin{figure*}[!ht]
\centering
\resizebox{0.95\linewidth}{!}{
\centering\includegraphics[scale=1.0]{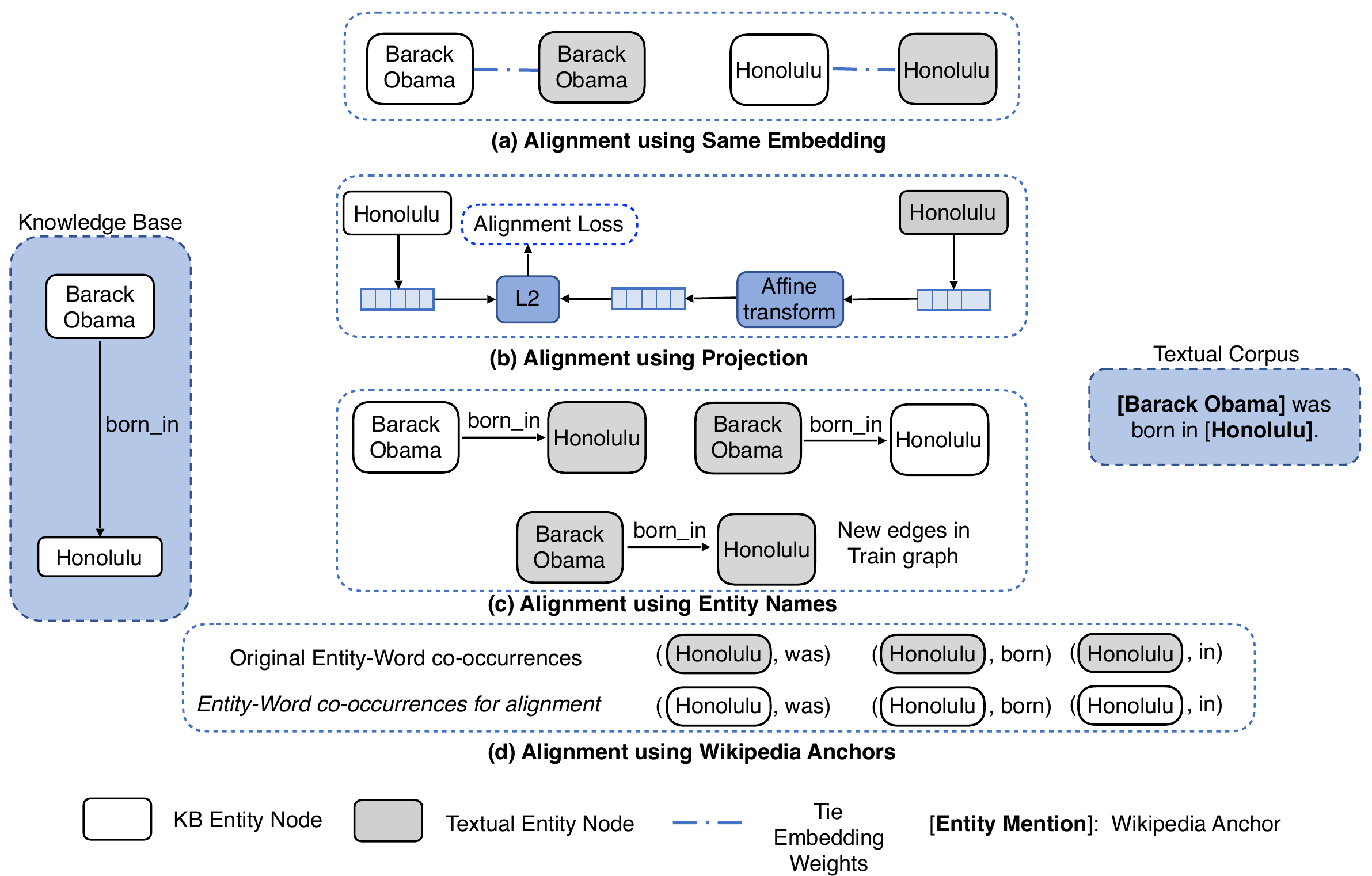}}
\caption{Schematic representation of different alignment methods\label{fig:alignment_fig}}
\end{figure*}

In this section, we describe the four alignment methods used in our study. At first, we describe the component models used in all alignment methods - the KB embedding model and the skip-gram model.\\



\subsection{Knowledge Base embedding model}
We use the TransE model \cite{bordes2013translating} to learn the KB embeddings.
We use the loss function proposed in \citet{sun2019rotate} as our KB embedding objective.
\begin{equation*}
\begin{aligned}[b]
&\mathcal{L}_{KB} =\\
&\sum_{(\bm{h},\bm{r},\bm{t}) \in S\cup S^{'}} \textrm{log}(1+\textrm{exp}(y * (-\gamma + d_r(\bm{h}, \bm{t}))))
\end{aligned}
\end{equation*}
Here, $d_r(\bm{h}, \bm{t}) = \norm{\bm{h}+\bm{r}-\bm{t}}_2$ denotes the score function for the triple $(h, r, t)$, $S$ denotes the set of positive triples and $S^{'}$ denotes the set of corrupted triples obtained by replacing the head or tail of a positive triple with a random entity. $\gamma$ is a hyper-parameter which denotes the margin and $y$ denotes the label (+1 for positive triple and -1 for negative triple).

\subsection{Skip-gram model}
The skip-gram model learns the embeddings of words and entities by modeling the word-word, word-entity and entity-entity co-occurrences. We use the skip-gram model proposed in \citet{yamada2016joint} for learning the word and entity representations. Let $\mathcal{W}$
and $\mathcal{E}$ denote the set of all words and entities in the vocabulary respectively and $c$ denote the size of the context window.
\begin{itemize}
    \item \textbf{Word-Word co-occurrence model}: The skip-gram model is trained to predict the target word given a context word. Given a sequence of $N$ words $w_1, w_2, \cdots, w_N$, the skip-gram model maximizes the following objective:
    $$ \mathcal{L}_{ww} = \sum_{n=1}^N \sum_{-c\leq j \leq c;j\neq 0} \textrm{log}\;P(w_{n+j}|w_n)$$ where
   $ p(w_O|w_I) = \frac{exp({v^{'}_{w_I}}^T v_{w_O})}{\sum_{w \in \mathcal{W}} exp({v^{'}_{w_I}}^T v_{w})}$. Here, $v^{'}_w$ and $v_w$ denote the input and output representations of the word $w$ respectively. The input representations are used as the final representations for both words and entities.
    \item \textbf{Word-Entity co-occurrence model}:
In the word-entity co-occurrence model, the model is trained to predict the context words of an entity pointed to by the target anchor. The training objective corresponding to the word-entity co-occurrences is
\begin{align*}
\mathcal{L}_{we} =\sum_{(e_i, C_{e_i})\in \mathcal{A}}\sum_{w_o\in C_{e_i}}\textrm{log}\;p(w_o|e_i)
\end{align*}
Here, $\mathcal{A}$ denotes the set of anchors in the corpus. Each anchor consists of an entity $e_i$ and its context words (represented by $C_{e_i}$).
The conditional probability $p(w_o|e_i)$ is given by:
$$ p(w_O|e_i) = \frac{exp({v^{'}_{e_i}}^T v_{w_O})}{\sum_{w \in \mathcal{W}} exp({v^{'}_{e_i}}^T v_{w})}$$

    \item \textbf{Entity-Entity co-occurrence model}:
The entity-entity co-occurrence model learns to predict incoming links of an entity (denoted by $C_e$) given an entity $e$.
$$ \mathcal{L}_{ee} =\sum_{e_i \in E}\sum_{e_o\in C_{e_i};e_i\neq e_o}\textrm{log}\;p(e_o|e_i)$$
$$ p(e_O|e_i) = \frac{exp({v^{'}_{e_i}}^T v_{e_O})}{\sum_{e \in \mathcal{E}} exp({v^{'}_{e_i}}^T v_{e})}$$
\end{itemize}
In practice, the probabilities involved in the skip-gram model are estimated using negative sampling \cite{mikolov2013distributed}. The overall objective is the sum of the three objectives for each type of co-occurrence.
$$ \mathcal{L}_{SG} = \mathcal{L}_{ww} + \mathcal{L}_{we} + \mathcal{L}_{ee}$$

\subsection{Alignment methods}
We align the entity pairs in KB and text corpus using a set of seed entity pairs, which are obtained from a mapping between Wikidata and Wikipedia. This mapping is constructed from the metadata associated with the Wikidata entities. The set of entities present in the TransE model and the skip-gram model is denoted by $\mathcal{E}_{TE}$ and $\mathcal{E}_{SG}$ respectively. 

\begin{enumerate}[label=(\alph*)]

\item \textbf{Alignment using same embedding}:
In this approach, we use the same embedding for the shared entities in the KB and text corpus. There is no separate alignment loss for this method.

\item \textbf{Alignment using Projection}:
Inspired by the multilingual word embedding approaches \cite{mikolov2013exploiting, faruqui2014improving} which use a linear transformation to map word embeddings from one space to another, we use an affine transformation from the skip-gram vector space to the TransE vector space to align the entity representations.

The alignment loss is calculated as a squared L2 norm between the transformed skip-gram entity embeddings and the corresponding TransE entity embeddings. The vectors $\bm{e}_{TE}$ and $\bm{e}_{SG}$ denote the TransE and skip-gram versions of embeddings of the entity $\bm{e}$ respectively.
\begin{equation*}
\begin{aligned}[b]
&\mathcal{L}_{align} =\\
&\sum_{\bm{e} \in \mathcal{E}_{SG}\cap \mathcal{E}_{TE}}\norm{(\bm{W} \bm{e}_{SG} + \bm{b}) - \bm{e}_{TE}}_2^{2}
\end{aligned}
\end{equation*}

\item \textbf{Alignment using Entity Names}:
In this alignment technique inspired by \citet{wang2014knowledge}, for a particular triple $(h, r, t)$ in the KB, if an equivalent entity $e_h$ exists in the text corpus, we add an additional triple $(e_h, r, t)$ to the KB. Similarly, if an equivalent entity $e_t$ also exists for the entity $t$, we add the triples $(h, r, e_t)$ and $(e_h, r, e_t)$ to the KB. The term ``name graph'' is used to denote this subgraph of additional triples.
\begin{equation*}
\begin{aligned}[b]
&\mathcal{L}_{align} =\\
&\sum_{(h, r, t)\; \in \textrm{ KB}} \mathbbm{1}_{[h \in \mathcal{E}_{SG} \land t \in \mathcal{E}_{SG}]} d_r(\bm{w}_h, \bm{w}_t)+ \\
&\mathbbm{1}_{[t \in \mathcal{E}_{SG}]}d_r(\bm{h}, \bm{w}_t) +\mathbbm{1}_{[h \in \mathcal{E}_{SG}]}d_r(\bm{w}_h, \bm{t})
\end{aligned}
\end{equation*}


\item \textbf{Alignment using Wikipedia Anchors}
This alignment technique is motivated by a similar technique proposed in \citet{wang2014knowledge}. Here, we introduce an alignment loss term in which for word-entity co-occurrences, we substitute the textual entity embedding by its KB counterpart in the skip-gram objective. Let $\bm{e}^i_{te}$ denote the embedding of the KB entity equivalent to the textual entity $e_i$.
\begin{equation*}
\begin{aligned}[b]
&\mathcal{L}_{align} =\\
&\sum_{(e_i, C_{e_i})\in \mathcal{A}}\sum_{w_o\in C_{e_i}} \textrm{log}\;\sigma(exp( {\bm{e}^i_{te}}^T {v_{w_O}}))+\\
&\sum_{i=1}^k \mathbb{E}_{w_i \sim P_n(\mathcal{W})} [\textrm{log}\;\sigma(-exp({\bm{e}^i_{te}}^T {v_{w_i}}))]
\end{aligned}
\end{equation*}
Here, $P_n(\mathcal{W})$ denotes the noise distribution over words and $k$ is the number of negative samples.
\end{enumerate}

The final objective for training these models becomes
$$ \mathcal{L} = \mathcal{L}_{KB} + \mathcal{L}_{SG} + \lambda \mathcal{L}_{align}$$
Here, $\lambda$ denotes the balance parameter which controls the extent of influence of alignment on the embeddings of each of the individual vector spaces.
An illustration of the different alignment methods used in our study is given in Figure~\ref{fig:alignment_fig}.

%% file: dataset.tex
We use Wikipedia as the text corpus and Wikidata \cite{vrandevcic2014wikidata} as the knowledge base.
We use the Wikidata version dated 16 December 2020 and the Wikipedia version dated 3 December 2020 for all of our experiments. The term \textit{support set} (as used in the subsequent sections), denoted by $\mathcal{S}$, is used to refer to the intersection set of Wikidata entities and entities in Wikipedia for which an article is present. 

\noindent \textbf{Dataset preprocessing.} We pre-process the original set of Wikidata triples and filter out entities and relations with frequency less than 10 and 5 respectively. This results in a KB with 14.64 M entities, 1222 relations, and 261 M facts. Similarly, we preprocess Wikipedia and filter out words from the vocabulary with frequency less than 10. However, we utilize the entire entity set of Wikipedia to maximize the size of the support set. After processing, the Wikipedia vocabulary consists of 2.1 M words and 12.3 M entities.

%% file: experiments.tex
\subsection{Experimental Setup} \label{sec:expt_setup}
We compare the performance of different alignment methods using two evaluation tasks - \textbf{few-shot link prediction} and \textbf{analogical reasoning}. The few-shot link prediction task is designed to test the capability of the alignment model to inject the relational information present in text into the knowledge base embeddings. The train-test set for this task is constructed such that the test set contains triples corresponding to a subset of entities in the support set, but each of these entities is observed only once in the training triples set. Thus, the model is tasked to do link prediction on entities that occur rarely in the training set (hence the term ``few-shot''). The training and test sets consist of 260.1 M and 110.8 K triples respectively. For this setting, both entities of each triple in the test set are contained in the support set.

The purpose of the analogical reasoning task is to test the information flow from the knowledge-base embeddings to the skip-gram embeddings. This task was first proposed in \citet{mikolov2013distributed} to test the syntactic and semantic information present in learned word embeddings. We choose the top 50 relations from the set of one-to-one and many-to-one relations based on the frequency of occurrence and construct a dataset of 1000 analogical reasoning examples for each relation. The 1st pair of entities is randomly chosen from the training triples set, as the pair of entities involved in that relation. The 2nd pair of entities is obtained from the test triples set. 
More formally, given a pair of entities $(h_1, t_1)$ and the head entity of the 2nd pair $(h_2)$, the task is to predict the tail entity $(t_2)$ of the 2nd pair by comparing the cosine similarity between the embedding of candidate entity $(\bm{e}_{t_2})$ and $(\bm{e}_{h_2} + \bm{e}_{t_1} - \bm{e}_{h_1})$.

\noindent \textbf{Evaluation protocol.}
For link prediction evaluation on a given test triple $(h, r, t)$, we corrupt either the head entity (by generating triplets like $(h^{'}, r, t)$) or the tail entity (by generating triplets like $(h, r, t^{'})$) of the triple and then rank the score of correct entity amongst all entities in the candidate set. Due to the extremely large entity vocabulary size in Wikidata, we restrict the size of the candidate set to a sample of 1000 entities whose types lie in the set of permissible domain/range types for that relation \cite{lerer2019pytorch, krompass2015type}. In cases where the number of such entities is less than 1000, we choose the entire set of those entities. In addition, we filter any positive triplets (triplets that exist in the KB) from the set of negative triplets for this evaluation, also known as \textit{filtered evaluation} setting. We report results on standard evaluation metrics - Mean Rank (MR), Hits@1, and Hits@10. For this task, we compare the TransE model and the KB-side embeddings of different alignment methods.

For the analogical reasoning task, we report Mean Rank (MR), Hits@1, and Hits@10 by ranking the correct entity $t_2$ against the entities in the candidate set. The candidate set for the tail entity $t_2$ is a set of 1K entities sampled from the support set (excluding $h_1$, $h_2$ and $t_1$) according to the node degree. All reported metrics are macro-averaged over the results for different relations. Here, we compare the skip-gram model embeddings with the textual embeddings obtained from different alignment methods.

\subsection{Implementation}
The scale of the training data (both the Wikidata Knowledge Base and the Wikipedia corpus) is huge, so the efficient implementation of the model is a key challenge. For efficient implementation of the TransE model, we used the DGL-KE \cite{zheng2020dgl} library. It uses graph partitioning to train across multiple partitions of the knowledge base in parallel and incorporates engineering optimizations like efficient negative sampling to reduce the training time by orders of magnitude compared to naive implementations. The skip-gram model is implemented using PyTorch \cite{paszke2019pytorch} and Wikipedia2vec \cite{yamada2020wikipedia2vec} libraries.

For training, we optimize the parameters of the TransE and skip-gram models alternately in each epoch. We use the Adagrad \cite{duchi2011adaptive} optimizer for the KBE model and SGD for the skip-gram model. For both models, the training is done by multiple processes asynchronously using the Hogwild \cite{niu2011hogwild} approach. This introduces additional challenges like synchronizing the weights of parameters among different training processes. We choose the values of balance parameter for each of the two evaluation tasks based on the performance of aligned KB and textual embeddings on a small set of analogy examples (disjoint from the analogy test set used in the main evaluation).
Our implementation can serve as a good resource to do a similar large-scale analysis of KB-Text alignment approaches in the future.

\input{overall_results_table.tex}

\subsection{Overall Results}
The overall results for the two evaluation tasks are given in Table~\ref{tab:results_overall}. For the few-shot link prediction task, we observe that all the alignment techniques lead to improved performance of the KB embeddings over the naive TransE baseline. The Same Embedding alignment approach performs the best followed by Entity Name alignment, Projection, and alignment using Wikipedia Anchors. The use of the same embeddings for the shared entities helps in propagating the factual knowledge present in the text to the KB more efficiently, so the Same Embedding alignment performs better than others. The Entity Name alignment approach is worse than the Same embedding alignment approach since the test set entities occur less often in the train set (as the dataset is few-shot). So, the name graph doesn't make a substantial difference here.  

For the analogical reasoning task, the results show that all alignment approaches obtain an improvement over the naive skip-gram baseline. The Entity Name alignment approach performs the best followed by Projection, Same Embedding alignment, and alignment using Wikipedia Anchors. The good performance of the Entity Name alignment approach could be explained by the fact that for every test analogy example $(e_{h_1}, e_{t_1}, e_{h_2}, e_{t_2})$, there is a relation $r$ present between the entity pairs $(e_{h_1}, e_{t_1})$ and $(e_{h_2}, e_{t_2})$, although that is unobserved. Since $e_h$ and $e_t$ also occur in the KB, due to the extra added triples, the KB reasoning process incorporates the relation $r$ in these embeddings, just like it does for KB entities $h$ and $t$. 
The other approaches viz. Same Embedding alignment, Projection, and Wikipedia Anchors don't have a mechanism for explicit KB reasoning like the Entity Name alignment approach. The Projection technique outperforms the Same Embedding alignment as the embeddings in the two spaces are less tightly coupled in the former, so it can take advantage of the complementary relational information in textual as well as the KB embeddings.

\subsection{Fine-grained Analysis}
In this section, we present a fine-grained analysis of the efficacy of the alignment methods w.r.t. changes in training data size and whether the test set entities belong to the support set. We also study the impact of balance parameter on the performance of the two evaluation tasks. Due to resource constraint, we do this analysis on two representative methods of different nature - Projection alignment and Same Embedding alignment.

\noindent \textbf{Effect of Training data size.} To study and differentiate the impact of entities present in the support set on the performance of the few-shot link prediction task, we create two versions of the training set with different sizes:
\begin{enumerate}[label=(\alph*)]
\item \textit{Full version}: In this version of the training set, we include all triples in Wikidata which don't violate the few-shot property of the dataset. This is the same as the training set for the evaluation proposed in Section~\ref{sec:expt_setup}. 
\item \textit{Support version}: In this version of the training set, we exclude triples from the \textit{full} version whose either head or tail entity isn't present in the support set.
\end{enumerate}

Next, we try to analyze the impact of whether the head/tail entity of the test triple is present in the support set $\mathcal{S}$, on the few-shot link prediction performance.
To this end, we create two versions of test sets:
\begin{enumerate}[label=(\alph*)]
\item  \textit{Both in support}: Both head and tail entity of the triple lie in the support set.
\item \textit{Missing support}: Atleast one out of the head/tail entity of the triple doesn't lie in the support set.
\end{enumerate}

The statistics for this dataset are given in Table~\ref{tab:dataset_stat}.

\input{overall_balance_param_tbl.tex}

\input{stats_tbl.tex}

The results for the training data size analysis for different alignment methods on Test set (Both in support) are shown in Table~\ref{tab:few_shot_size_setting}.
The results show that for both Projection and Same Embedding alignment approach, the performance is significantly better with using the full training set of triples instead of just the support set. This shows that triples involving non-support set entities play a vital role in helping learn better entity and relation representations which in turn helps in injecting textual information to the KB embeddings via alignment.

\noindent \textbf{Effect of Support set for Test triples.}
Here, we investigate the performance of the few-shot link prediction task for triples whose entities may not lie in the support set.
The results for this evaluation are given in Table~\ref{tab:few_shot_missing_support}. We observe that there is no significant gain in performance for any of the alignment methods over the simple TransE baseline. This shows these alignment methods are only effective for triples whose both entities lie in the support set. 

\input{train_size_ablation_tbl.tex}

\noindent \textbf{Effect of balance parameter.}
In this analysis, we study the role of balance parameter for the Projection alignment method. This parameter controls the extent of alignment between the two embedding spaces. The higher the value of the balance parameter, the more the embedding tries to capture the entity information from the other embedding space, rather than its own. The results of this study are shown in Table~\ref{tab:overall_balance_param}. The peak performance for the few-shot link prediction task is obtained for balance parameter = 1e0 in terms of Hits@1 and Hits@10. Whereas, for the analogical reasoning task, the peak performance is obtained for balance parameter = 1e-3. This difference in the optimal value of the balance parameter can be explained by the fact that the skip-gram objective relies on cosine similarity which is more sensitive to changes in the values of vector embeddings than the TransE model. We show this analytically. Let $(h, r, t)$ be a KB triple and let $\bm{h}$, $\bm{r}$, and $\bm{t}$ denote the embeddings of $h$, $r$, and $t$ respectively.
The partial derivative of score function of the triple w.r.t. $\bm{h}$ is given by
\begin{align*}
    d_r(\bm{h}, \bm{t}) &= \norm{\bm{h}+\bm{r}-\bm{t}}_2 \\
    \norm{\frac{\partial d_r(\bm{h}, \bm{t})}{\partial \bm{h}}}_2 &= \norm{\frac{(\bm{h} + \bm{r} - \bm{t})}{\norm{\bm{h}+\bm{r}-\bm{t}}_2}}_2 = 1
\end{align*}
Similarly, let $(u, v)$ be an entity-word pair in the text corpus. Let $\bm{u}$ and $\bm{v}$ denote the embeddings of $u$ and $v$ respectively.
The partial derivative of the score function for the entity-word pair $(u, v)$ w.r.t. $\bm{u}$ is given by
\begin{align*}
    d(\bm{u}, \bm{v}) &= exp(\bm{u}^T \bm{v}) \\
    \norm{\frac{\partial d(\bm{u}, \bm{v})}{\partial \bm{u}}}_2 &= \norm{(\bm{u}^T \bm{v}) \bm{v}}_2 = (\bm{u}^T \bm{v}) \norm{\bm{v}}_2
\end{align*}
The value of $\norm{\frac{\partial d_r(\bm{h}, \bm{t})}{\partial \bm{h}}}_2$ equals 1 whereas for the skip-gram model,  $\norm{\frac{\partial d(\bm{u}, \bm{v})}{\partial \bm{u}}}_2 = (\bm{u}^T \bm{v}) \norm{\bm{v}}_2$ which is greater than 1, as seen empirically. This shows that the skip-gram embeddings are more sensitive to delta changes in values of the parameters. For them to be reasonably assigned with their KB counterparts without losing the textual information, thus a lower value of balance parameter is optimal.

\input{covid_tbl.tex}

\subsection{Case study on COVID related triples}
Recently, the COVID pandemic \cite{fauci2020covid} has been responsible for bringing a tremendous change in the lives of people across the globe. Through this case study, we demonstrate that aligning embedding representations can help us do knowledge base completion for recent events like COVID-19. We selected 4 relevant relations (``Risk factor'', ``Symptoms'', ``Medical Condition'' and ``Cause of Death'') with atleast 10 triples in the difference between March 2020 and December 2020 snapshots of Wikidata. We use the March 2020 Wikidata and December 2020 Wikipedia to train the alignment models and do link prediction on these triples. For each of the relations, we keep the COVID-19 entity (Entity ID: Q84263196) unchanged and corrupt the other entity in the triple. This would correspond to asking questions like ``What are the symptoms of COVID-19?'', ``Who died due to COVID-19?'' etc. The results are shown in Table~\ref{tab:covid_results_mr}.

We observe that the Projection model obtains a decent improvement over the TransE model on the link prediction task on these triples in terms of Mean Rank. Similarly, the Same Embedding alignment model obtains outperforms the TransE baseline for three out of four relations. This case study gives a real-life use-case of how the text information can be injected into the KB embeddings using alignment in scenarios when such information is not yet curated in the KB in structured form.

%% file: overall_results_table.tex
\begin{table*}[htbp]
\centering
\small
\begin{tabular}{lcccccc}
\toprule
& \multicolumn{3}{c}{\bfseries Few-shot Link Prediction} & \multicolumn{3}{c}{\bfseries Analogical Reasoning} \\ \cmidrule{2-7}
\bfseries Model                                  & \bfseries MR        & \bfseries Hits@1    & \bfseries Hits@10   & \bfseries MR      & \bfseries Hits@1   & \bfseries Hits@10   \\ \midrule
TransE                            & 187      & 20.3     & 40.4     &    --     &     --    &      --     \\
Skip-gram &    --     &     --    &    -- & 25	& 50.6	& 78.0 \\
\midrule
Projection                        & 134        & 22.9     & 47.2      & 12    & 65.9    & 89.0     \\
\specialcell{Same Embedding align.}    & \textbf{102}     & \textbf{30.7}     & \textbf{51.8}     &  11	& 60.7	& 87.5       \\
\specialcell{Entity Name align.\\} & 116    & 23.1     & 46.7     &  \textbf{8}	& \textbf{66.5}	& \textbf{91.0} \\
\specialcell{Wikipedia Anchors align.} & 138    & 25.8     & 46.2      &  14 &	56.1 &	84.8     \\ \bottomrule
\end{tabular}
\caption{Overall results for both evaluation tasks.}
\label{tab:results_overall}
\end{table*}

%% file: overall_balance_param_tbl.tex
\begin{table*}[!htbp]
\centering
\small
\begin{tabular}{lcccccc}
\toprule
& \multicolumn{3}{c}{\bfseries Few-shot Link Prediction}  & \multicolumn{3}{c}{\bfseries Analogical Reasoning}         \\ \cmidrule{2-7}
\bfseries Model           & \bfseries MR           & \bfseries Hits@1         & \bfseries Hits@10       & \bfseries MR             & \bfseries Hits@1         & \bfseries Hits@10        \\ \midrule
TransE          & 187       & 20.3          & 40.4         &  --              &  --              &  --              \\
Skip-gram       & --             &  --              &  --             & 25          & 50.6          & 78.0             \\
\midrule
Projection (balance param.=1e-4) & 188       & 20.4          & 40.4         & 14          & 65.0          & 88.0          \\
Projection (balance param.=1e-3) & 186       & 20.5          & 40.5         & 12          & \textbf{65.9} & \textbf{89.0} \\
Projection (balance param.=1e-2) & 184       & 20.6          & 40.6         & \textbf{10} & 61.4          & 87.3          \\
Projection (balance param.=1e-1) & 169       & 20.7          & 42.0         & 16          & 57.8          & 84.2          \\
Projection (balance param.=1e0)  & 134       & \textbf{22.9} & \textbf{47.2} & 23          & 49.6          & 78.9          \\ 
Projection (balance param.=1e1)  & \textbf{129} & 21.4          & 43.1         &  26              &  42.2              & 75.4              \\ \bottomrule
\end{tabular}
\caption{Overall results for Projection alignment model for different values of balance parameter.}
\label{tab:overall_balance_param}
\end{table*}

%% file: stats_tbl.tex
\begin{table}[htbp]
\centering
\small
\begin{tabular}{ll}
\toprule
\textbf{Dataset}                       & \textbf{No. of triples} \\ \midrule
Train set (Full)              & 260.1 M       \\
Train set (Support)           & 17.1 M        \\
Test set (Both in support)    & 110.8 K        \\
Test set (Missing support) & 38.3 K         \\ \bottomrule
\end{tabular}
\caption{Few-shot Link Prediction dataset statistics.}
\label{tab:dataset_stat}
\end{table}

%% file: train_size_ablation_tbl.tex
\begin{table}[!ht]
\centering
\small
\begin{tabular}{lc}
\toprule
\textbf{Model}                         & \textbf{Mean Rank}  \\ \midrule
\specialcell{Projection (Full)}    & 134                   \\
\specialcell{Projection (Support)} & 208                   \\ 
\midrule
\specialcell{Same embed. align. (Full)}    &  102       \\
\specialcell{Same embed. align. (Support)} & 184        \\ 
\midrule
\specialcell{TransE (Full)}            & 188               \\
\specialcell{TransE (Support)}         & 255               \\ \bottomrule                  
\end{tabular}
\caption{Results for different training set sizes for Few-Shot Link Prediction task.}
\label{tab:few_shot_size_setting}
\end{table}

\begin{table}[!ht]
\centering
\small
\begin{tabular}{lc}
\toprule
\textbf{Model}                          & \textbf{Mean Rank} \\ \midrule
\specialcell{Projection (Full)}    & 208                     \\
\specialcell{Same embed. align. (Full)}         &  207                   \\
\specialcell{TransE (Full)}             & 213                     \\ \bottomrule                  
\end{tabular}
\caption{Results for Missing Support Test Set (Few-shot Link Prediction task).}
\label{tab:few_shot_missing_support}
\end{table}

%% file: covid_tbl.tex
\begin{table}[!htbp]
\centering
\small
\resizebox{.45\textwidth}{!}{
\begin{tabular}{lccc}
\toprule
\textbf{Relation}                & \textbf{TransE} & \textbf{Projection} & \specialcell{\textbf{Same Embed.}} \\ \midrule
Risk factor             & 312 & 261 &  \textbf{153}  \\
Symptoms                & 37  & \textbf{36}  &  39  \\
\specialcell{Medical cond.}       & 371 & \textbf{267} &  330  \\
Cause of death          & 314 & \textbf{246} &  299  \\ \bottomrule 
\end{tabular}
}
\caption{Link Prediction results for COVID-19 case study (Mean Rank).}
\label{tab:covid_results_mr}
\end{table}

%% file: conclusion.tex
In this work, we presented a systematic study of different alignment approaches that can be applied to align entity representations in a knowledge base and textual corpora. By evaluating on the few-shot link prediction task and analogical reasoning task, we found that although all approaches have the desired outcome, i.e., to incorporate information from the other modality, some approaches perform better than others on a particular task. We also analyzed the impact of different factors such as the size of the training set, the presence of test set entities in the support set, and the balance parameter on the evaluation task performance. 
We believe our evaluation framework, as well as jointly trained embeddings can serve as a useful resource for future research and applications.

%% file: acknowledgements.tex
We would like to thank the anonymous reviewers for their helpful comments. This research was sponsored by a gift grant from Fujitsu and the Ohio Supercomputer Center \cite{OhioSupercomputerCenter1987}.

%% file: acl2021.bbl
\begin{thebibliography}{46}
\expandafter\ifx\csname natexlab\endcsname\relax\def\natexlab#1{#1}\fi

\bibitem[{Auer et~al.(2007)Auer, Bizer, Kobilarov, Lehmann, Cyganiak, and
  Ives}]{auer2007dbpedia}
S{\"o}ren Auer, Christian Bizer, Georgi Kobilarov, Jens Lehmann, Richard
  Cyganiak, and Zachary Ives. 2007.
\newblock Dbpedia: A nucleus for a web of open data.
\newblock In \emph{The semantic web}, pages 722--735. Springer.

\bibitem[{Bollacker et~al.(2007)Bollacker, Tufts, Pierce, and
  Cook}]{bollacker2007platform}
Kurt Bollacker, Patrick Tufts, Tomi Pierce, and Robert Cook. 2007.
\newblock A platform for scalable, collaborative, structured information
  integration.
\newblock In \emph{Intl. Workshop on Information Integration on the Web
  (IIWeb’07)}, pages 22--27.

\bibitem[{Bordes et~al.(2013)Bordes, Usunier, Garcia-Duran, Weston, and
  Yakhnenko}]{bordes2013translating}
Antoine Bordes, Nicolas Usunier, Alberto Garcia-Duran, Jason Weston, and Oksana
  Yakhnenko. 2013.
\newblock Translating embeddings for modeling multi-relational data.
\newblock In \emph{Advances in neural information processing systems}, pages
  2787--2795.

\bibitem[{Cao et~al.(2017)Cao, Huang, Ji, Chen, and Li}]{cao2017bridge}
Yixin Cao, Lifu Huang, Heng Ji, Xu~Chen, and Juanzi Li. 2017.
\newblock Bridge text and knowledge by learning multi-prototype entity mention
  embedding.
\newblock In \emph{Proceedings of the 55th Annual Meeting of the Association
  for Computational Linguistics (Volume 1: Long Papers)}, pages 1623--1633.

\bibitem[{Castells et~al.(2007)Castells, Fern{\'a}ndez, and
  Vallet}]{Castells2007AnAO}
P.~Castells, M.~Fern{\'a}ndez, and D.~Vallet. 2007.
\newblock An adaptation of the vector-space model for ontology-based
  information retrieval.
\newblock \emph{IEEE Transactions on Knowledge and Data Engineering}, 19.

\bibitem[{Center(1987)}]{OhioSupercomputerCenter1987}
Ohio~Supercomputer Center. 1987.
\newblock \href {http://osc.edu/ark:/19495/f5s1ph73} {Ohio supercomputer
  center}.

\bibitem[{Chen et~al.(2017)Chen, Tian, Yang, and
  Zaniolo}]{chen2016multilingual}
Muhao Chen, Yingtao Tian, Mohan Yang, and Carlo Zaniolo. 2017.
\newblock Multilingual knowledge graph embeddings for cross-lingual knowledge
  alignment.
\newblock In \emph{Proceedings of the 26th International Joint Conference on
  Artificial Intelligence}, IJCAI'17, page 1511–1517. AAAI Press.

\bibitem[{Dettmers et~al.(2018)Dettmers, Minervini, Stenetorp, and
  Riedel}]{dettmers2018convolutional}
Tim Dettmers, Pasquale Minervini, Pontus Stenetorp, and Sebastian Riedel. 2018.
\newblock Convolutional 2d knowledge graph embeddings.
\newblock In \emph{Thirty-Second AAAI Conference on Artificial Intelligence}.

\bibitem[{Dong et~al.(2014)Dong, Gabrilovich, Heitz, Horn, Lao, Murphy,
  Strohmann, Sun, and Zhang}]{dong2014knowledge}
Xin Dong, Evgeniy Gabrilovich, Geremy Heitz, Wilko Horn, Ni~Lao, Kevin Murphy,
  Thomas Strohmann, Shaohua Sun, and Wei Zhang. 2014.
\newblock Knowledge vault: A web-scale approach to probabilistic knowledge
  fusion.
\newblock In \emph{Proceedings of the 20th ACM SIGKDD international conference
  on Knowledge discovery and data mining}, pages 601--610.

\bibitem[{Duchi et~al.(2011)Duchi, Hazan, and Singer}]{duchi2011adaptive}
John Duchi, Elad Hazan, and Yoram Singer. 2011.
\newblock Adaptive subgradient methods for online learning and stochastic
  optimization.
\newblock \emph{Journal of machine learning research}, 12(7).

\bibitem[{Faruqui and Dyer(2014)}]{faruqui2014improving}
Manaal Faruqui and Chris Dyer. 2014.
\newblock Improving vector space word representations using multilingual
  correlation.
\newblock In \emph{Proceedings of the 14th Conference of the European Chapter
  of the Association for Computational Linguistics}, pages 462--471.

\bibitem[{Fauci et~al.(2020)Fauci, Lane, and Redfield}]{fauci2020covid}
Anthony~S Fauci, H~Clifford Lane, and Robert~R Redfield. 2020.
\newblock Covid-19—navigating the uncharted.

\bibitem[{Gardner et~al.(2014)Gardner, Talukdar, Krishnamurthy, and
  Mitchell}]{gardner2014incorporating}
Matt Gardner, Partha Talukdar, Jayant Krishnamurthy, and Tom Mitchell. 2014.
\newblock Incorporating vector space similarity in random walk inference over
  knowledge bases.
\newblock In \emph{Proceedings of the 2014 conference on empirical methods in
  natural language processing (EMNLP)}, pages 397--406.

\bibitem[{Gladkova et~al.(2016)Gladkova, Drozd, and
  Matsuoka}]{gladkova2016analogy}
Anna Gladkova, Aleksandr Drozd, and Satoshi Matsuoka. 2016.
\newblock \href {https://doi.org/10.18653/v1/N16-2002} {Analogy-based detection
  of morphological and semantic relations with word embeddings: what works and
  what doesn{'}t.}
\newblock In \emph{Proceedings of the {NAACL} Student Research Workshop}, pages
  8--15, San Diego, California. Association for Computational Linguistics.

\bibitem[{Hao et~al.(2016)Hao, Zhang, He, Liu, and Zhao}]{hao2016joint}
Yanchao Hao, Yuanzhe Zhang, Shizhu He, Kang Liu, and Jun Zhao. 2016.
\newblock A joint embedding method for entity alignment of knowledge bases.
\newblock In \emph{China Conference on Knowledge Graph and Semantic Computing},
  pages 3--14. Springer.

\bibitem[{Ji et~al.(2015)Ji, He, Xu, Liu, and Zhao}]{ji2015knowledge}
Guoliang Ji, Shizhu He, Liheng Xu, Kang Liu, and Jun Zhao. 2015.
\newblock Knowledge graph embedding via dynamic mapping matrix.
\newblock In \emph{Proceedings of the 53rd Annual Meeting of the Association
  for Computational Linguistics and the 7th International Joint Conference on
  Natural Language Processing (Volume 1: Long Papers)}, pages 687--696.

\bibitem[{Krompa{\ss} et~al.(2015)Krompa{\ss}, Baier, and
  Tresp}]{krompass2015type}
Denis Krompa{\ss}, Stephan Baier, and Volker Tresp. 2015.
\newblock Type-constrained representation learning in knowledge graphs.
\newblock In \emph{International semantic web conference}, pages 640--655.
  Springer.

\bibitem[{Lao and Cohen(2010)}]{lao2010relational}
Ni~Lao and William~W Cohen. 2010.
\newblock Relational retrieval using a combination of path-constrained random
  walks.
\newblock \emph{Machine learning}, 81(1):53--67.

\bibitem[{Lao et~al.(2012)Lao, Subramanya, Pereira, and Cohen}]{lao2012reading}
Ni~Lao, Amarnag Subramanya, Fernando Pereira, and William Cohen. 2012.
\newblock Reading the web with learned syntactic-semantic inference rules.
\newblock In \emph{Proceedings of the 2012 Joint Conference on Empirical
  Methods in Natural Language Processing and Computational Natural Language
  Learning}, pages 1017--1026.

\bibitem[{Lerer et~al.(2019)Lerer, Wu, Shen, Lacroix, Wehrstedt, Bose, and
  Peysakhovich}]{lerer2019pytorch}
Adam Lerer, Ledell Wu, Jiajun Shen, Timothee Lacroix, Luca Wehrstedt, Abhijit
  Bose, and Alex Peysakhovich. 2019.
\newblock {PyTorch-BigGraph: A Large-scale Graph Embedding System}.
\newblock In \emph{Proceedings of the 2nd SysML Conference}, Palo Alto, CA,
  USA.

\bibitem[{Lin et~al.(2015)Lin, Liu, Sun, Liu, and Zhu}]{lin2015learning}
Yankai Lin, Zhiyuan Liu, Maosong Sun, Yang Liu, and Xuan Zhu. 2015.
\newblock Learning entity and relation embeddings for knowledge graph
  completion.
\newblock In \emph{Twenty-ninth AAAI conference on artificial intelligence}.

\bibitem[{Mikolov et~al.(2013{\natexlab{a}})Mikolov, Le, and
  Sutskever}]{mikolov2013exploiting}
Tomas Mikolov, Quoc~V Le, and Ilya Sutskever. 2013{\natexlab{a}}.
\newblock Exploiting similarities among languages for machine translation.
\newblock \emph{arXiv preprint arXiv:1309.4168}.

\bibitem[{Mikolov et~al.(2013{\natexlab{b}})Mikolov, Sutskever, Chen, Corrado,
  and Dean}]{mikolov2013distributed}
Tomas Mikolov, Ilya Sutskever, Kai Chen, Greg~S Corrado, and Jeff Dean.
  2013{\natexlab{b}}.
\newblock Distributed representations of words and phrases and their
  compositionality.
\newblock In \emph{Advances in neural information processing systems}, pages
  3111--3119.

\bibitem[{Nathani et~al.(2019)Nathani, Chauhan, Sharma, and
  Kaul}]{nathani2019learning}
Deepak Nathani, Jatin Chauhan, Charu Sharma, and Manohar Kaul. 2019.
\newblock \href {https://doi.org/10.18653/v1/P19-1466} {Learning
  attention-based embeddings for relation prediction in knowledge graphs}.
\newblock In \emph{Proceedings of the 57th Annual Meeting of the Association
  for Computational Linguistics}, pages 4710--4723, Florence, Italy.
  Association for Computational Linguistics.

\bibitem[{Nguyen et~al.(2018)Nguyen, Nguyen, Nguyen, and
  Phung}]{nguyen2017novel}
Dai~Quoc Nguyen, Tu~Dinh Nguyen, Dat~Quoc Nguyen, and Dinh Phung. 2018.
\newblock \href {https://doi.org/10.18653/v1/N18-2053} {A novel embedding model
  for knowledge base completion based on convolutional neural network}.
\newblock In \emph{Proceedings of the 2018 Conference of the North {A}merican
  Chapter of the Association for Computational Linguistics: Human Language
  Technologies, Volume 2 (Short Papers)}, pages 327--333, New Orleans,
  Louisiana. Association for Computational Linguistics.

\bibitem[{Niu et~al.(2011)Niu, Recht, Re, and Wright}]{niu2011hogwild}
Feng Niu, Benjamin Recht, Christopher Re, and Stephen~J. Wright. 2011.
\newblock Hogwild! a lock-free approach to parallelizing stochastic gradient
  descent.
\newblock In \emph{Proceedings of the 24th International Conference on Neural
  Information Processing Systems}, NIPS'11, page 693–701, Red Hook, NY, USA.

\bibitem[{Paszke et~al.(2019)Paszke, Gross, Massa, Lerer, Bradbury, Chanan,
  Killeen, Lin, Gimelshein, Antiga et~al.}]{paszke2019pytorch}
Adam Paszke, Sam Gross, Francisco Massa, Adam Lerer, James Bradbury, Gregory
  Chanan, Trevor Killeen, Zeming Lin, Natalia Gimelshein, Luca Antiga, et~al.
  2019.
\newblock Pytorch: An imperative style, high-performance deep learning library.
\newblock In \emph{Advances in neural information processing systems}, pages
  8026--8037.

\bibitem[{Pennington et~al.(2014)Pennington, Socher, and
  Manning}]{pennington2014glove}
Jeffrey Pennington, Richard Socher, and Christopher~D Manning. 2014.
\newblock Glove: Global vectors for word representation.
\newblock In \emph{Proceedings of the 2014 conference on empirical methods in
  natural language processing (EMNLP)}, pages 1532--1543.

\bibitem[{Schlichtkrull et~al.(2018)Schlichtkrull, Kipf, Bloem, Van Den~Berg,
  Titov, and Welling}]{schlichtkrull2018modeling}
Michael Schlichtkrull, Thomas~N Kipf, Peter Bloem, Rianne Van Den~Berg, Ivan
  Titov, and Max Welling. 2018.
\newblock Modeling relational data with graph convolutional networks.
\newblock In \emph{European Semantic Web Conference}, pages 593--607. Springer.

\bibitem[{Shang et~al.(2019)Shang, Tang, Huang, Bi, He, and
  Zhou}]{shang2019end}
Chao Shang, Yun Tang, Jing Huang, Jinbo Bi, Xiaodong He, and Bowen Zhou. 2019.
\newblock End-to-end structure-aware convolutional networks for knowledge base
  completion.
\newblock In \emph{Proceedings of the AAAI Conference on Artificial
  Intelligence}, volume~33, pages 3060--3067.

\bibitem[{Shen et~al.(2015)Shen, Wang, and Han}]{Shen2015EntityLW}
Wei Shen, Jianyong Wang, and Jiawei Han. 2015.
\newblock Entity linking with a knowledge base: Issues, techniques, and
  solutions.
\newblock \emph{IEEE Transactions on Knowledge and Data Engineering},
  27:443--460.

\bibitem[{Socher et~al.(2013)Socher, Chen, Manning, and
  Ng}]{socher2013reasoning}
Richard Socher, Danqi Chen, Christopher~D Manning, and Andrew Ng. 2013.
\newblock Reasoning with neural tensor networks for knowledge base completion.
\newblock In \emph{Advances in neural information processing systems}, pages
  926--934.

\bibitem[{Suchanek et~al.(2007)Suchanek, Kasneci, and
  Weikum}]{suchanek2007yago}
Fabian~M Suchanek, Gjergji Kasneci, and Gerhard Weikum. 2007.
\newblock Yago: a core of semantic knowledge.
\newblock In \emph{Proceedings of the 16th international conference on World
  Wide Web}, pages 697--706.

\bibitem[{Sun et~al.(2019)Sun, Deng, Nie, and Tang}]{sun2019rotate}
Zhiqing Sun, Zhi-Hong Deng, Jian-Yun Nie, and Jian Tang. 2019.
\newblock \href {https://openreview.net/forum?id=HkgEQnRqYQ} {Rotate: Knowledge
  graph embedding by relational rotation in complex space}.
\newblock In \emph{International Conference on Learning Representations}.

\bibitem[{Toutanova et~al.(2015)Toutanova, Chen, Pantel, Poon, Choudhury, and
  Gamon}]{toutanova2015representing}
Kristina Toutanova, Danqi Chen, Patrick Pantel, Hoifung Poon, Pallavi
  Choudhury, and Michael Gamon. 2015.
\newblock Representing text for joint embedding of text and knowledge bases.
\newblock In \emph{Proceedings of the 2015 conference on empirical methods in
  natural language processing}, pages 1499--1509.

\bibitem[{Vrande{\v{c}}i{\'c} and Kr{\"o}tzsch(2014)}]{vrandevcic2014wikidata}
Denny Vrande{\v{c}}i{\'c} and Markus Kr{\"o}tzsch. 2014.
\newblock Wikidata: a free collaborative knowledgebase.
\newblock \emph{Communications of the ACM}, 57(10):78--85.

\bibitem[{Wang et~al.(2014)Wang, Zhang, Feng, and Chen}]{wang2014knowledge}
Zhen Wang, Jianwen Zhang, Jianlin Feng, and Zheng Chen. 2014.
\newblock Knowledge graph and text jointly embedding.
\newblock In \emph{Proceedings of the 2014 conference on empirical methods in
  natural language processing (EMNLP)}, pages 1591--1601.

\bibitem[{Wang et~al.(2018)Wang, Lv, Lan, and Zhang}]{wang2018cross}
Zhichun Wang, Qingsong Lv, Xiaohan Lan, and Yu~Zhang. 2018.
\newblock Cross-lingual knowledge graph alignment via graph convolutional
  networks.
\newblock In \emph{Proceedings of the 2018 Conference on Empirical Methods in
  Natural Language Processing}, pages 349--357.

\bibitem[{Xie et~al.(2016)Xie, Liu, Jia, Luan, and Sun}]{xie2016representation}
Ruobing Xie, Zhiyuan Liu, Jia Jia, Huanbo Luan, and Maosong Sun. 2016.
\newblock Representation learning of knowledge graphs with entity descriptions.
\newblock In \emph{Proceedings of the AAAI Conference on Artificial
  Intelligence}.

\bibitem[{Yamada et~al.(2020)Yamada, Asai, Sakuma, Shindo, Takeda, Takefuji,
  and Matsumoto}]{yamada2020wikipedia2vec}
Ikuya Yamada, Akari Asai, Jin Sakuma, Hiroyuki Shindo, Hideaki Takeda,
  Yoshiyasu Takefuji, and Yuji Matsumoto. 2020.
\newblock {W}ikipedia2{V}ec: An efficient toolkit for learning and visualizing
  the embeddings of words and entities from {W}ikipedia.
\newblock In \emph{Proceedings of the 2020 Conference on Empirical Methods in
  Natural Language Processing: System Demonstrations}, pages 23--30.
  Association for Computational Linguistics.

\bibitem[{Yamada et~al.(2016)Yamada, Shindo, Takeda, and
  Takefuji}]{yamada2016joint}
Ikuya Yamada, Hiroyuki Shindo, Hideaki Takeda, and Yoshiyasu Takefuji. 2016.
\newblock \href {https://doi.org/10.18653/v1/K16-1025} {Joint learning of the
  embedding of words and entities for named entity disambiguation}.
\newblock In \emph{Proceedings of The 20th {SIGNLL} Conference on Computational
  Natural Language Learning}, pages 250--259, Berlin, Germany. Association for
  Computational Linguistics.

\bibitem[{Yao and Van~Durme(2014)}]{Yao2014InformationEO}
Xuchen Yao and Benjamin Van~Durme. 2014.
\newblock \href {https://doi.org/10.3115/v1/P14-1090} {Information extraction
  over structured data: Question answering with {F}reebase}.
\newblock In \emph{Proceedings of the 52nd Annual Meeting of the Association
  for Computational Linguistics (Volume 1: Long Papers)}, pages 956--966,
  Baltimore, Maryland. Association for Computational Linguistics.

\bibitem[{Yu et~al.(2017)Yu, Yin, Hasan, dos Santos, Xiang, and
  Zhou}]{Yu2017ImprovedNR}
Mo~Yu, Wenpeng Yin, Kazi~Saidul Hasan, Cicero dos Santos, Bing Xiang, and Bowen
  Zhou. 2017.
\newblock \href {https://doi.org/10.18653/v1/P17-1053} {Improved neural
  relation detection for knowledge base question answering}.
\newblock In \emph{Proceedings of the 55th Annual Meeting of the Association
  for Computational Linguistics (Volume 1: Long Papers)}, pages 571--581,
  Vancouver, Canada. Association for Computational Linguistics.

\bibitem[{Zheng et~al.(2020{\natexlab{a}})Zheng, Song, Ma, Tan, Ye, Dong,
  Xiong, Zhang, and Karypis}]{zheng2020dgl}
Da~Zheng, Xiang Song, Chao Ma, Zeyuan Tan, Zihao Ye, Jin Dong, Hao Xiong, Zheng
  Zhang, and George Karypis. 2020{\natexlab{a}}.
\newblock Dgl-ke: Training knowledge graph embeddings at scale.
\newblock In \emph{Proceedings of the 43rd International ACM SIGIR Conference
  on Research and Development in Information Retrieval}, SIGIR '20, page
  739–748, New York, NY, USA. Association for Computing Machinery.

\bibitem[{Zheng et~al.(2020{\natexlab{b}})Zheng, Rao, Song, Zhang, Xiao, Fang,
  Yang, and Niu}]{Zheng2020PharmKGAD}
Shuangjia Zheng, J.~Rao, Y.~Song, Jixian Zhang, Xianglu Xiao, E.~Fang, Yuedong
  Yang, and Zhangming Niu. 2020{\natexlab{b}}.
\newblock Pharmkg: a dedicated knowledge graph benchmark for bomedical data
  mining.
\newblock \emph{Briefings in bioinformatics}.

\bibitem[{Zhong et~al.(2015)Zhong, Zhang, Wang, Wan, and
  Chen}]{zhong2015aligning}
Huaping Zhong, Jianwen Zhang, Zhen Wang, Hai Wan, and Zheng Chen. 2015.
\newblock Aligning knowledge and text embeddings by entity descriptions.
\newblock In \emph{Proceedings of the 2015 Conference on Empirical Methods in
  Natural Language Processing}, pages 267--272.

\end{thebibliography}
